\documentclass{article}
\usepackage{spconf,amsmath,graphicx}
\usepackage{enumitem}
\setlist{nosep, leftmargin=14pt}

\usepackage{mwe}
\usepackage{graphicx}
\usepackage{subcaption}
\usepackage{url}
\usepackage{float}

\title{MCI Detection using fMRI time series embeddings of Recurrence plots  }

 \name{Ninad Aithal$^{\star}$ \qquad Chakka Sai Pradeep $^{\dagger}$ \qquad Neelam Sinha$^{\star}$}

\address{$^{\star}$Centre for Brain Research, Indian Institute of Science, Bengaluru \\
         $^{\dagger}$International Institute of Information Technology, Bengaluru}

\begin{document}

\maketitle
% --------------------

\begin{abstract} The human brain can be conceptualized as a dynamical system. Utilizing  resting state fMRI time series imaging, we can study the underlying dynamics at ear-marked Regions of Interest (ROIs) to understand structure or lack thereof. This differential behavior could be key to understanding the neurodegeneration and also to classify between healthy and Mild Cognitive Impairment (MCI) subjects. In this study, we consider 6 brain networks spanning over 160 ROIs derived from Dosenbach template, where each network consists of 25-30 ROIs. Recurrence plot, extensively used to understand evolution of time series, is employed. Representative time series at each ROI is converted to its corresponding recurrence plot visualization, which is subsequently condensed to low-dimensional feature embeddings through Autoencoders.
The performance of the proposed method is shown on fMRI volumes of 100 subjects (balanced data), taken from publicly available ADNI dataset. Results obtained show peak classification accuracy of 93\% among the 6 brain networks, mean accuracy of 89.3\% thereby illustrating promise in the proposed approach.

\end{abstract}
% --------------------

% --------------------
% --------------------

% --------------------
\section{Introduction}
\label{sec:intro}

Neurodegeneration remains a significant healthcare challenge for which effective solutions are yet to be discovered. Diseases like Alzheimer's (AD), without a known cure, present formidable hurdles. Mild Cognitive Impairment (MCI) is an intermediary stage before neurodegeneration that can lead to AD. It is well-known that without intervention AD can be life-crippling. Hence detecting mci early offers a crucial window for treatment, allowing a higher chance of recovery. Understanding/identifying MCI is imperative towards this. In this work, we utilize fMR time series to identify and understand MCI.

Several existing methods convert time series to images in order to analyze and understand the underlying dynamics.  Popular methods include GAF, MTF, cellular automata, etc. that transform time series into images, based on diverse properties such as Alternate co-ordinate systems and Transition probabilites\cite{csai}. However, a tool that exploits non-linear analysis is Recurrence plot visualization, which attempts to understand the dynamics that drive the time series. Recurrence plots \cite{recurrence} have gained popularity and are used across diverse domains, from financial markets for uncovering hidden market patterns to neuroscience, where they aid in understanding connectivity. Their versatility and utility make recurrence plots a popular tool. Recurrence plots are capable of revealing signatures of predictability, stationarity, fractal behavior, chaos and noise. Inferences using recurrence plots traditionally used Recurrence Quantitification Analysis (RQA) methods like Recurrence rate ($RR$), Determinism ($DET$), Laminarity ($LAM$) and Mean diagonal line length ($L$). However, in our approach, we utilize Recurrence plots by condensing them into low-dimensional latent space embeddings.

% --------------------

% --------------------
\section{Literature review}
\label{sec:lit}
The biggest challenge in fMRI data analysis is that it is very high dimensional. It has been studied over the years, in various ways such as Generalized Linear Model (GLM) \cite{glm}, Multi Voxel Pattern Analysis (MVPA) \cite{MVPA}, non-linear analysis \cite{nla} and most recently, using deep learning \cite{deep} including the idea of transformers \cite{transformer}. The ability to combine different frameworks to leverage the abilities of data-driven deep learning approaches has shown immense promise.

Several reported works in literature utilize concepts of non-linear time series analysis followed up with CNN-based frameworks for tasks such as detection, classification, quantification, etc. A recent work \cite{hatami} reported the usage of recurrence plots and employed CNN-based classifier for time series classification. Yet another work \cite{rpandgraph} reports the usage of cross recurrence plots and graph analysis to study brain dynamics from fMRI data. The authors utilized RQA features such as Recurrence rate (RR), Determinism (DET), and maximal length of vertical lines. A recent work \cite{crqa} utilizing cross recurrence quantification analysis on resting state fMRI time series reported its efficacy in detecting changes in neuropsychiatric lupus. Yet another recent work  \cite{kangrqa} employed RQA to measure the duration, predictability, and complexity of the periodic processes of the time series. 
% --------------------
\begin{figure*}[ht]
\centering
\includegraphics[width=\linewidth]{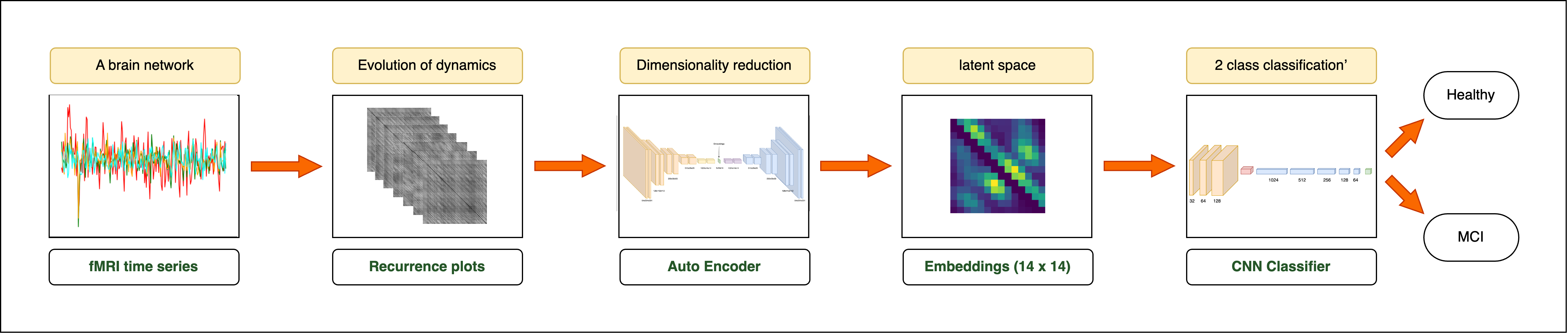}
\caption{Overview of Proposed methodology}
\label{fig:workflow}
\end{figure*}

% --------------------
\section{Proposed Methodology}
\label{sec:methodology}

% --------------------
Our method aims to classify between healthy and MCI subjects across 6 brain networks using recurrence plots and their latent space embeddings.The contributions of the paper are:
\begin{itemize}
    \item Proposed MCI classification using latent space embeddings of recurrence plots.   
    \item Encoding information across all ROIs of a brain network into a single 2D representation, used here for classification. 
\end{itemize} 
The block diagram in Fig.\ref{fig:workflow} shows an overview of the proposed method. All codes and results obtained are available at github link \url{ https://tinyurl.com/ycyhka6m}.

\subsection{fMRI Time series }
\label{ssec:subhead}
% \vspace{0.2cm}
In this study, we utilize a balanced dataset of fMRI volume time series of 50 MCI and 50 Healthy subjects taken from the publicly avaialble ADNI \cite{adni} dataset. In order to extract the time series,  structural divisions of the cerebral cortex\cite{dosenbach}, as provided in Dosenbach atlas, consisting of 160 ROIs across 6 brain networks, are used. All 6 networks, namely cingulo-opercular, frontoparietal, default mode, sensorimotor, occipital, and cerebellar networks are studied. Each ROI is represented by a sphere with a 5 mm radius. Within each of these regions, a representative time series, obtained using Regional Homogenity (ReHo) processing is extracted. 

In order to utilize non-linear analysis, the optimal values of the required parameters, embedding dimension ($M$), time lag ($\tau$),  and number of states ($K$), are determined using the traditional Cao's algorithm \cite{cao}. For time series $\{v_1, v_2, \ldots, v_n\}$ consisting of $N$ timestamps, we create $K$ state vectors. Each state vector $\overrightarrow{s_i}$ is an $M$-dimensional vector defined as, 

\begin{equation}
   \overrightarrow{s_i} = (v_i, v_{i+\tau}, v_{i+2\tau}, \ldots, v_{i+(N-1)\tau})
   \label{eq:1}
\end{equation}

\subsection{Data Matrix }
\label{ssec:dm}
% \vspace{0.2cm}

Data matrix, represented as D, consists of these states, stacked together, as shown in Eq. \ref{eq:2}. The dimension of $D$ is, $(K \times M)$. This matrix D will now be utilized to define the Recurrence matrix, RM.

\begin{equation}
   D = 
   \begin{bmatrix}
   \overrightarrow{s_1}\\
   \overrightarrow{s_2}\\
   \vdots\\
   \overrightarrow{s_n}\\
    \end{bmatrix}_{(K \times M)}
    \label{eq:2}
\end{equation}

\subsection{Recurrence matrix (RM)}
\label{ssec:rm}
% \vspace{0.2cm}
Distance between two state vectors is computed using their Euclidean distance, quantifying the dis-similarity between them. Hence, for a given tuple (i,j), the distance between states, $\overrightarrow{s_i}$ and $\overrightarrow{s_j}$, is the $(i,j)^{th}$ element of the RM, given by Eq \ref{eq:3},

\begin{equation}
    RM(i)(j) = dist(\overrightarrow{s_i}, \overrightarrow{s_j})
    \label{eq:3}
\end{equation}

Hence, the resulting RM is as defined below:

\begin{equation}
    RM = 
    \begin{bmatrix}
    dist(\overrightarrow{s_1},\overrightarrow{s_1}) & \cdots & dist(\overrightarrow{s_1},\overrightarrow{s_K}) \\
    dist(\overrightarrow{s_2},\overrightarrow{s_1}) & \cdots & dist(\overrightarrow{s_2},\overrightarrow{s_K}) \\
    \vdots & \vdots & \vdots\\
    dist(\overrightarrow{s_K},\overrightarrow{s_1}) & \cdots  & dist(\overrightarrow{s_K},\overrightarrow{s_K}) \\
    \end{bmatrix}_{(K \times K)}    
    \label{eq:4}
\end{equation}

\begin{figure}[ht]
\centering
\includegraphics[width=\linewidth]{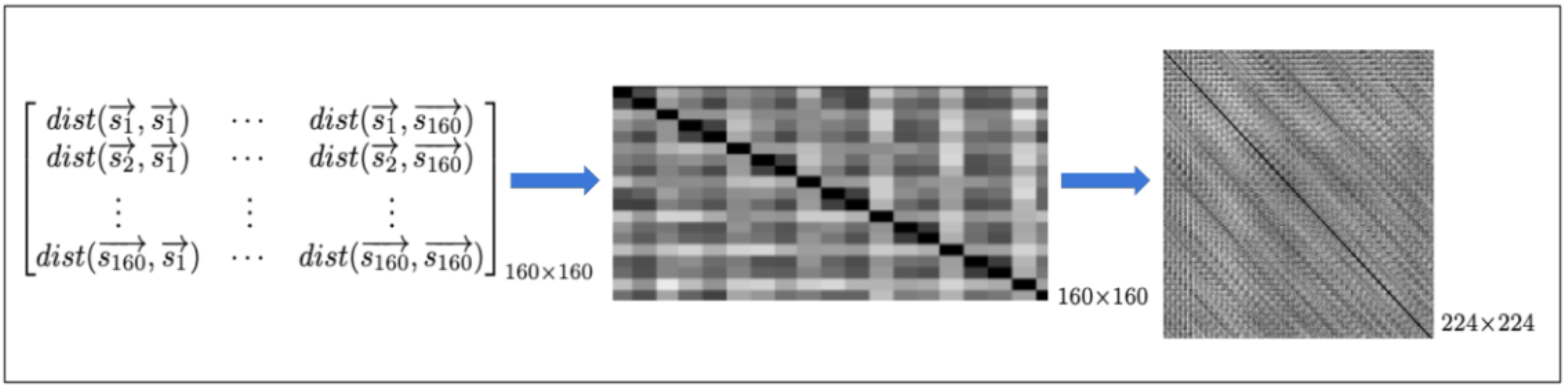}
\caption{Illustration of Conversion of an RM (for instance, of size 160 $\times$ 160, to uniformly-sized ($224\times224$) Recurrence Plot visualization.  
}
\label{fig:recurrencematrix}
\end{figure}

\subsubsection{Recurrence Plot Visualization }
\label{sssec:viz}
The Recurrence plot visualization shown in Fig. \ref{fig:recurrencematrix} is a gray-scale single channel image, capturing the evolution of the time series, as captured by the RM. Hence, the information at a specific ROI is now available as an image of size $1\times224\times224$. This leads to a high-dimensional representation of each brain network as a data point of Number of ROIs$\times224\times224$. In order to reduce the dimensionality and also to capture the essential characteristics, we utilize an autoencoder to obtain feature embeddings of size $14\times14$.

\subsection{Autoencoder}
\label{sssec:autoencoder}

\begin{figure}[ht]
\centering
\includegraphics[width=\linewidth]{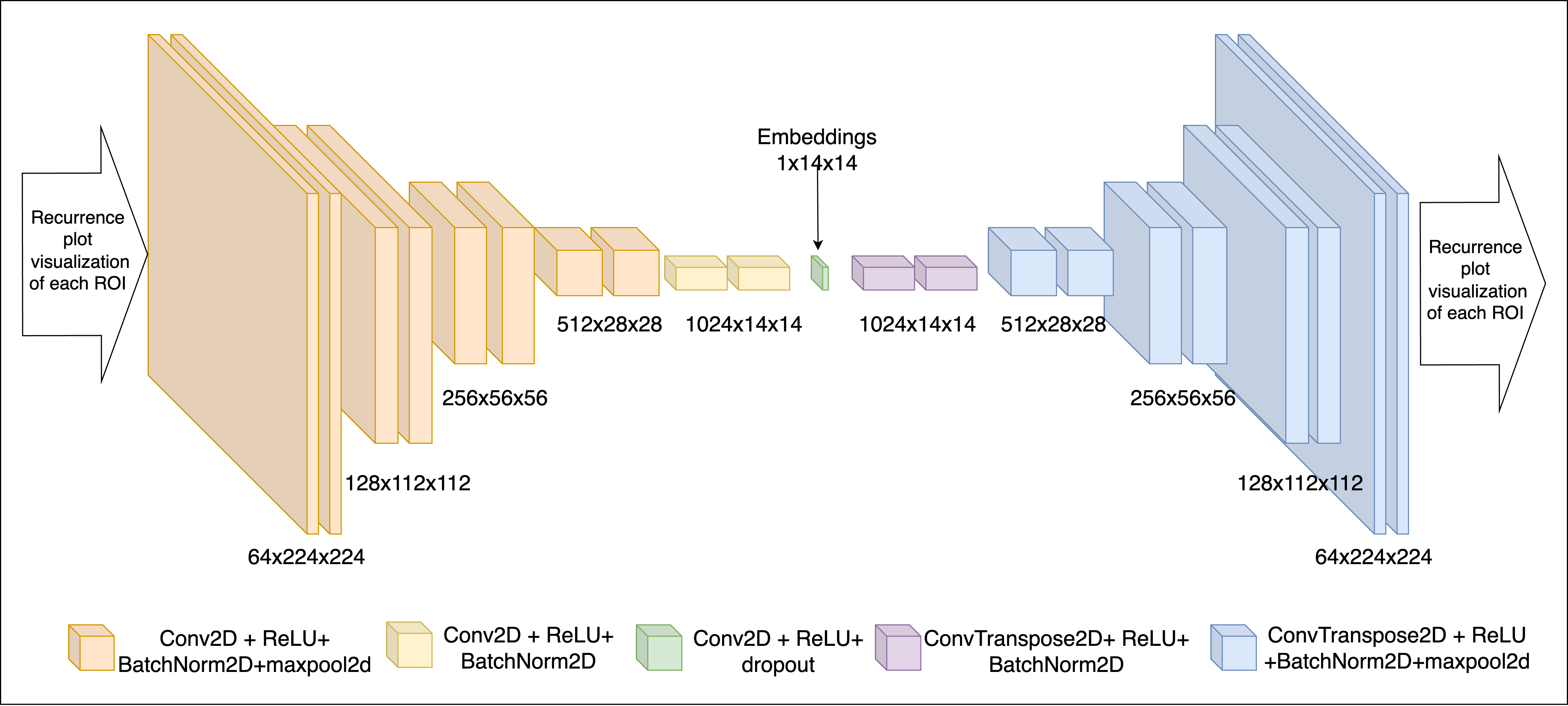}
\caption{Autoencoder architecture}
\label{fig:autoencoder}
\end{figure}

 CNN-based autoencoder, as shown in Fig. \ref{fig:autoencoder} is utilized to obtain the latent space embeddings of the recurrence plot visualizations derived from the time series. By design, all the networks are reduced to a $14\times14$ embedding (bottleneck). The input to the autoencoder comprises of the recurrence plot visualizations, making the input size, Number of ROIs $\times$ $224\times224$. The layers within the network capture the essential patterns in the Recurrence Plot Visualization. For training the autoencoder, the loss function consists of two terms, as shown in Eq.\ref{eq:5}. The first term is intended for reconstruction fidelity, while the second term penalizes the difference in Mean Structural Similarity Index (MSSIM) \cite{SSIM}. This ensures that the process preserves structural information. We use Adam optimizer for training. Autoencoder training was carried out on 2xT4 GPUs and took approximately 3 hours to train for 1000 epochs. A batch size of 16 and a train-test split of 0.2 were used in the training process.
 
\begin{equation}
L(x) = |x - \tilde{x}| + (1 - \text{MSSIM}(x, \tilde{x}))
\label{eq:5}
\end{equation}

% --------------------
% --------------------

\subsection{Classifier }
\label{ssec:classifier}
\begin{figure}[ht]
\centering
\includegraphics[width=\linewidth]{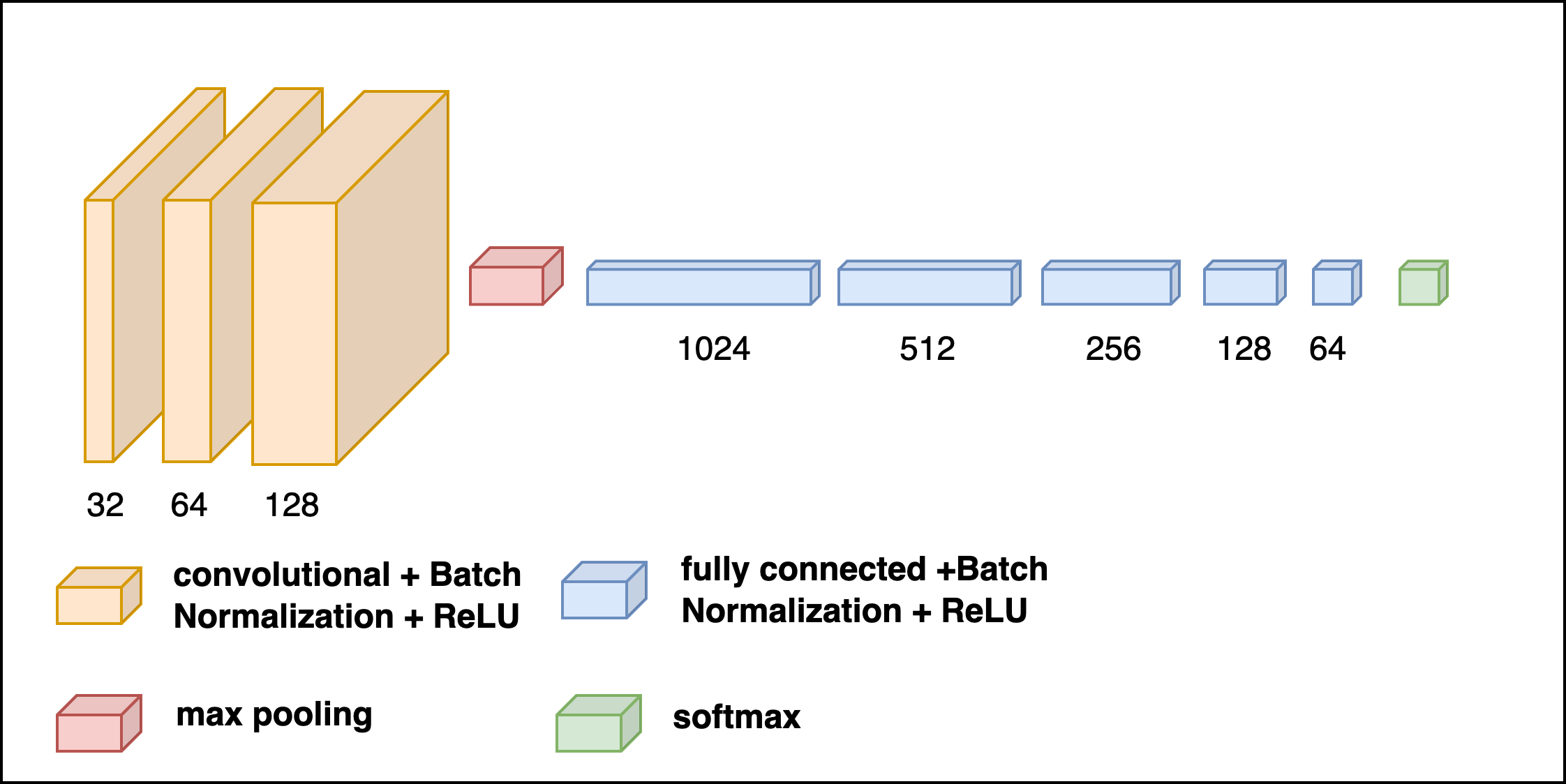}
\caption{CNN-based classifier architecture}
\label{fig:classifier}
\end{figure}

A custom CNN-based classifier as shown in Fig. \ref{fig:classifier}, is used. The input to the classifier is the set of latent space feature embeddings that are obtained from the Recurrence plot visualization of the time series. This network has 3 CNN-layers and 6 fully connected layers for prediction with Cross Entropy Loss function. Here again, Adam optimiser is used.

% --------------------
% --------------------

% --------------------
\section{Results}
\label{sec:results}
The proposed methodology requires fMRI timeseries as input. Each timeseries is converted to its corresponding Recurrence plot visualization, chosen here, to be of size $224\times224$. At this stage, each subject is represented by a high-dimensional input of following dimensions: (Number of ROIs $\times224\times224$). In order to compress and also to extract essential information, these images are now fed to an Autoencoder (whose loss function minimizes MSSIM difference) to obtain feature embeddings of size $14\times14$. This profiles each subject with a data point of dimensions only, $14\times14$. These feature embeddings corresponding to both classes of subjects, are now fed to a CNN-based custom classifier. The classification results obtained using the proposed approach are tabulated in Table \ref{table:Table1}. Peak classification accuracy of 93.3\% and mean accuracy (across all 6 brain networks) of 89.3\%, is obtained. The confusion matrix obtained for classifying latent space representation of ROIs of the DMN network is shown in Table\ref{table:Table2}.

\begin{table}[h]

\centering
\begin{tabular}{|l|c|c|c|c|}
\hline
\textbf{Network} & \textbf{Accuracy}& \textbf{Precision}& \textbf{Recall}\\
\hline
\textbf{Cerebellum}  & 93.33 \% & 0.89 & 1.00 \\
\textbf{Cingulo-Opercular} & 92.86\% & 1.00 & 0.83 \\
\textbf{Sensorimotor}& 92.86\% & 0.87 & 1.00 \\
\textbf{Default mode} & 92.31\% & 0.83 & 1.00 \\
\textbf{Frontoparietal} & 85.71\% & 1.00 & 0.80 \\
\textbf{Occipital}  & 78.57\% & 0.71 & 0.83 \\
\hline

\end{tabular}
\caption{Classification metrics of the model for healthy versus MCI using latent space embeddings of recurrence plots.}
\label{table:Table1}
\end{table}
\begin{table}[h]

\centering
\begin{tabular}{|l|c|c|c|}
\hline
\textbf{} & \textbf{Healthy}& \textbf{MCI}\\
\hline
\textbf{Healthy}  & 7  & 1 \\
\hline
\textbf{MCI} & 0 & 5 \\

\hline

\end{tabular}

\caption{Confusion matrix obtained using DMN network}
\label{table:Table2}
\end{table}

% --------------------

% --------------------
\begin{figure}[ht]

    \begin{subfigure}{0.225\textwidth}
        \centering
        \includegraphics[width=1.1\linewidth]{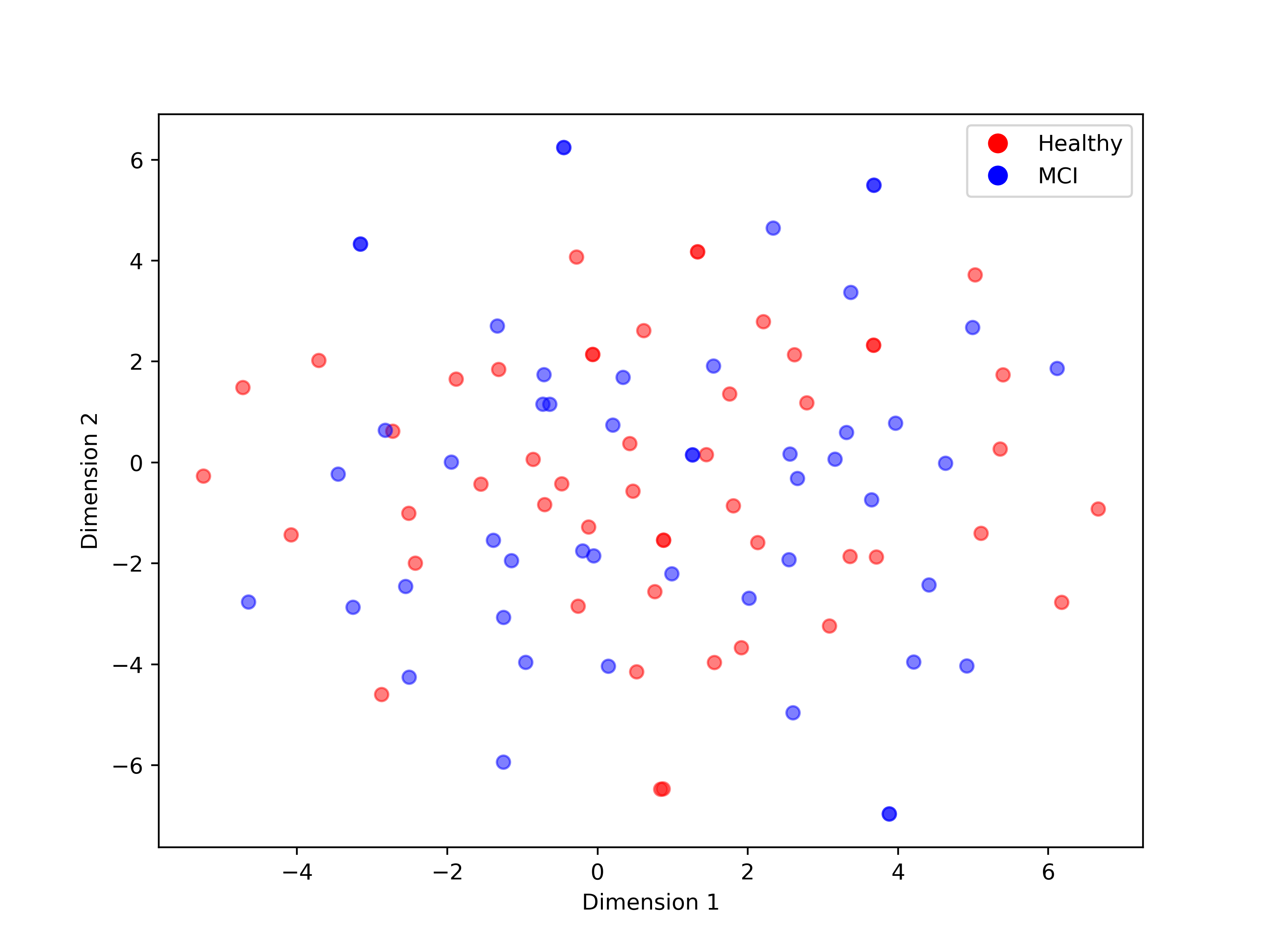}
        \caption{t-SNE visualization of RQA space}
    \end{subfigure}
    \hfill
    \begin{subfigure}{0.225\textwidth}
        \centering
        \includegraphics[width=\linewidth]{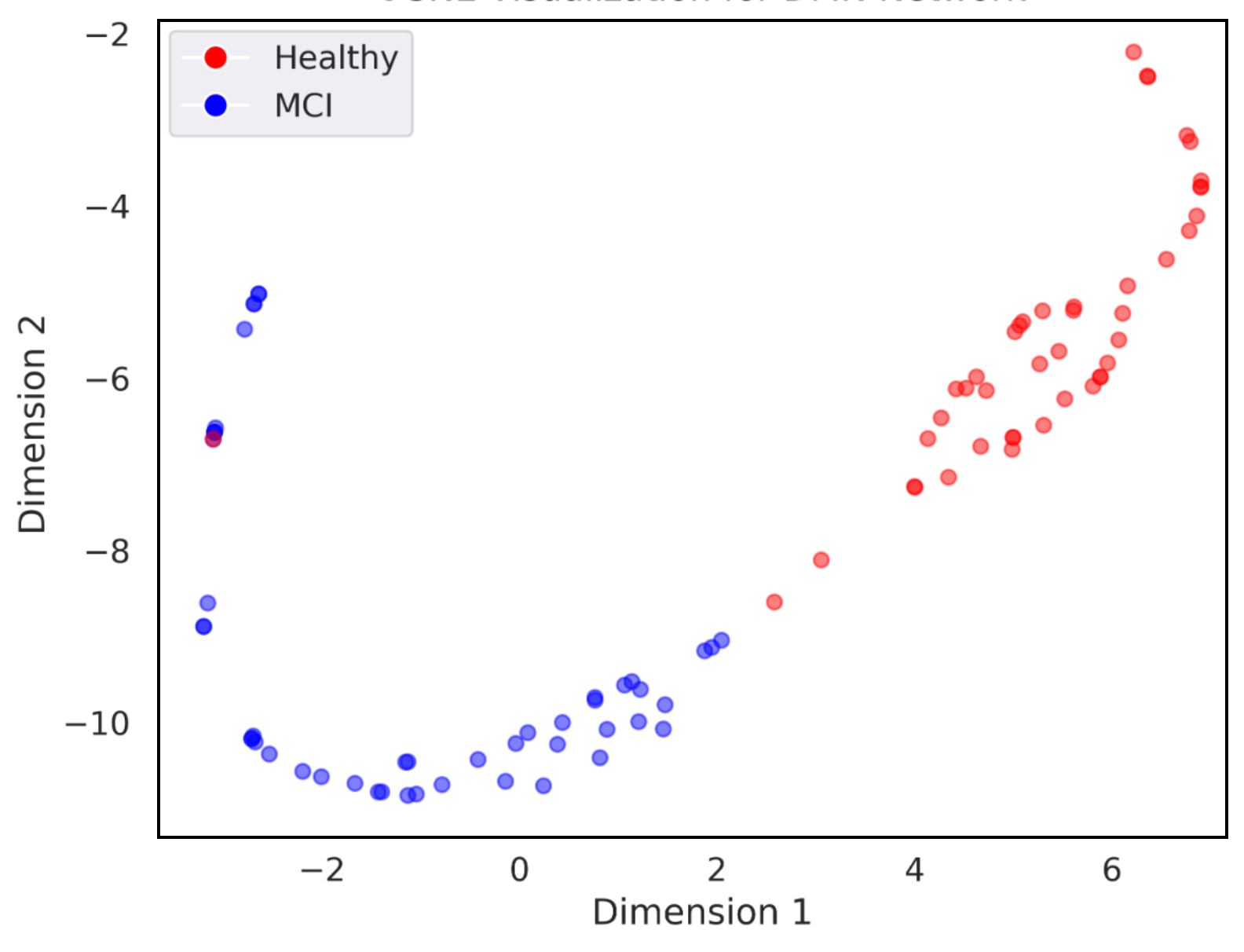}
        \caption{t-SNE visualization of latent space}
    \end{subfigure}
    \hfill
    \caption{Visualization comparison to illustrate inter-class separability achieved using the proposed method shown in (b), which remains a challenge for the traditional feature space using RQA shown in (a)}.
    \label{fig:comparetsne}
\end{figure}
% --------------------

% --------------------
\section{Discussion }
\label{sec:discussion}

Recurrence plots are information-rich and can be used to derive inferences such as structure, chaos, predictability or lack thereof. Traditionally these plots are processed using Recurrence Quantification Aalysis (RQA), which compute 13 different RQA measures as given in section \ref{sec:intro}. However, in our experiments, feature embeddings of Recurrence plots are utilized for classification. A comparison of the separability of feature space spanned by the obtained feature embeddings versus those spanned by RQA measures are shown in Figures \ref{fig:comparetsne}(a) and (b), respectively. Feature embeddings provide far better promise for classification compared to RQA measures. Comparison of classification performance, using the proposed latent space embeddings of recurrence plots for the 6 different brain networks, is tabulated in Table \ref{table:Table1}. Mean classification accuracy of 89.3\% and peak accuracy of 93.3\% is observed for Cerebellum network.

% --------------------

\section{Conclusion }
\label{sec:conclusion}

In this study, we propose the utility of non-linear time series analysis and latent space embeddings to detect MCI using fMRI data. The hypothesis is that the underlying dynamics at the affected brain regions are different in MCI as compared to healthy subjects, which is captured in the recurrence plot visualization. The information-rich high-dimensional representation is condensed to low-dimensional feature embeddings in the latent space, using autoencoders. This can be viewed as fusion of information across multiple ROIs, which can be also be processed for tasks like detection, clustering, representation, etc. Here, the disparities in feature embeddings are exploited for classification. Promising results of 93\% classification accuracy on publicly available ADNI data of 100 subjects (balanced dataset) illustrate the efficacy of the proposed approach.

\bibliographystyle{IEEEbib}
\bibliography{newref}

\end{document}